\documentclass[runningheads]{llncs}
\usepackage[T1]{fontenc}
\usepackage{graphicx}
\usepackage{multirow}
\usepackage{hyperref}
\hypersetup{
    colorlinks=true,
    linkcolor=blue,
    citecolor=blue,
    urlcolor=blue
}
\usepackage[nameinlink]{cleveref} 
\begin{document}
\title{MediRec: Enhancing Chinese Medication Recommendation with Explainable Clinical Reasoning}
\titlerunning{MediRec: Chinese Medication Recommendation}
%
\author{Juntao Li\inst{+} \and
Haobin Yuan\inst{+} \and
Ling Luo\inst{*} \and
Yuanyuan Sun \and
Jian Wang \and
Hongfei Lin}
\authorrunning{Li et al.}
%
\institute{College of Computer Science and Technology, Dalian University of Technology, Dalian 116024, China \\
\inst{*}To whom correspondence should be addressed: \email{lingluo@dlut.edu.cn}\\
\inst{+}Equal contribution}

\maketitle              
\begin{abstract}
Large language models (LLMs) have shown strong potential for clinical decision support through their advanced language understanding and reasoning capabilities. However, their application to Chinese clinical medication recommendation remains largely unexplored. Existing approaches are primarily developed on English electronic health record datasets and focus on coarse-grained medication code prediction, offering limited support for interpretable clinical decision-making. In this work, we propose MediRec, an explainable LLM-based framework for Chinese medication recommendation from electronic health records. MediRec combines clinically grounded reasoning-chain distillation with reinforcement learning to improve both recommendation accuracy and interpretability. Comprehensive experiments on a Chinese medication recommendation benchmark show that MediRec achieves strong performance, with an F1 score of 0.5813 and a Jaccard score of 0.4626. Further analyses indicate that MediRec generates clinically plausible recommendations with transparent reasoning, demonstrating its effectiveness for explainable medication decision support in Chinese healthcare settings.

\end{abstract}

\section{Introduction}

Medication recommendation is an important task in clinical decision support, aiming to suggest appropriate drug prescriptions based on a patient’s electronic health records. Accurate medication recommendation can help clinicians improve treatment efficiency, reduce prescription errors, and support safer personalized care. Owing to its practical clinical value, the development of accurate and reliable automated medication recommendation systems has attracted increasing attention in recent years.

With the rapid advancement of large language models (LLMs), several studies have begun to integrate them into medication recommendation tasks and have achieved promising results \cite{fan2026fine,liu2024large,zhao2025addressing}. However, existing LLM-based approaches still exhibit several critical limitations. First, most methods are built upon publicly available English MIMIC (Medical Information Mart for Intensive Care) databases \cite{johnson2023mimic,johnson2016mimic}, and typically use only partial patient information, such as diagnosis and procedures, to construct medication recommendation datasets. Second, current approaches predominantly focus on predicting coarse-grained drug categories rather than recommending specific fine-grained medications. As a result, these methods lack explicit clinical reasoning processes and offer limited interpretability for medication decisions. Furthermore, due to substantial differences in patient population demographics, disease distributions, and regional clinical guidelines, models trained on English datasets suffer from severe data distribution biases, limiting their adaptability to real-world clinical environments in China.

To address these challenges, we conduct a comprehensive study of LLM-based Chinese medication recommendation. We first benchmark several representative LLMs, including Llama \cite{grattafiori2024llama}, GLM \cite{glm2024chatglm}, Qwen \cite{qwen2025qwen25technicalreport}, and Baichuan \cite{dou2026baichuan}, under different prompting and fine-tuning strategies to analyze their capabilities and limitations in this task. We then propose MediRec, a two-stage explainable medication recommendation framework for Chinese electronic health records (EHRs). In the first stage, MediRec distills clinically grounded reasoning chains from patient records and drug instructions to provide cold-start supervision. In the second stage, the model is further optimized with reinforcement learning to improve both medication recommendation accuracy and reasoning quality. 

Our contributions are summarized as follows: First, we investigate fine-grained Chinese medication recommendation from EHRs and systematically benchmark representative LLMs under diverse prompting and fine-tuning strategies. Our results show that general-purpose LLMs struggle to perform this task reliably without domain-specific adaptation. Second, we propose MediRec, a two-stage explainable medication recommendation framework that distinctly combines clinical reasoning distillation with reinforcement learning optimization to elevate both recommendation accuracy and reasoning interpretability. Third, experimental results demonstrate that MediRec achieves strong performance on the Chinese discharge medication recommendation benchmark while generating more interpretable and clinically plausible medication recommendations.

\section{Related Work}

Medication recommendation aims to generate safe and effective prescriptions from patient medical records and has become an important clinical decision-support task \cite{wu2023dual,yang2021change,zhang2023hmfnet}. Driven by advancements in deep learning, neural network-based approaches have emerged as the dominant research paradigm in this domain. These methods are broadly categorized into two types: instance-based approaches and patient longitudinal history-based approaches.

Instance-based methods focus on a patient’s current medical condition \cite{gong2021smr,read2011classifier,zhang2017leap}. For example, LEAP \cite{zhang2017leap} builds label dependencies through a recurrent decoder and uses an attention mechanism to capture mapping relationships between diseases and medications. SMR \cite{gong2021smr} transforms the complex task of personalized safe medication recommendation as a graph-based link prediction problem by integrating multiple medical knowledge graphs and utilizing joint embedding learning. Longitudinal history-based methods leverage temporal dependencies across a patient’s historical records to predict future medication needs \cite{shang2019gamenet,wu2022conditional,yang2021safedrug}. For instance, GAMENet \cite{shang2019gamenet} integrates patient longitudinal health records and drug knowledge graphs, modeling drug combinations and drug–drug interactions through graph convolutional networks. COGNet \cite{wu2022conditional} introduces a copy-or-predict mechanism that effectively incorporates a patient's historical medication data to enhance both the accuracy and safety of recommendations.

More recently, LLMs have been applied to many downstream tasks. The development of efficient fine-tuning techniques \cite{dettmers2023qlora,hu2022lora} has significantly improved LLM performance in domain-specific scenarios \cite{zhang2025parameter}. In the medical domain, the Taiyi model \cite{luo2024taiyi} shows strong generalization ability across multilingual and multi-task settings by leveraging rich biomedical datasets alongside a two-stage fine-tuning strategy. Furthermore, Liu et al. \cite{liu2024large} provided an initial exploration into the application of LLMs for medication recommendation by distilling the parameters learned by fine-tuned LLMs into smaller, more efficient models. However, existing LLM-based approaches are primarily developed on English datasets and focus on coarse-grained medication prediction, offering limited support for Chinese clinical scenarios and interpretable reasoning. In contrast, our work investigates fine-grained Chinese medication recommendation and proposes an explainable LLM-based framework that improves both recommendation accuracy and interpretability.

\section{MediRec}
\begin{figure}
\includegraphics[width=\textwidth]{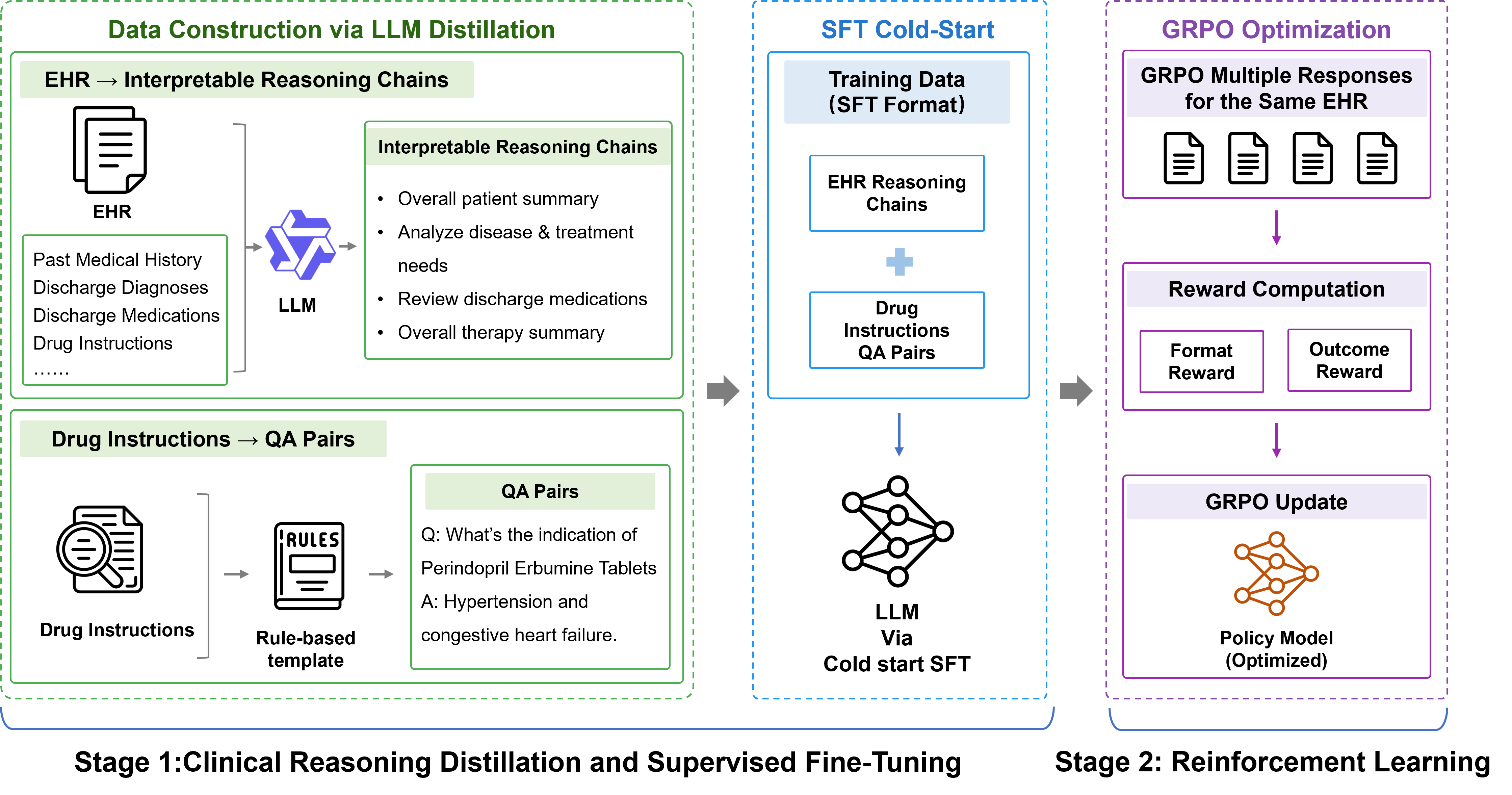}
\caption{Overview of our MediRec framework. Stage 1: A teacher LLM distills interpretable clinical reasoning chains from EHRs and drug instructions, which are converted into SFT training data; the model is then initialized through SFT. Stage 2: The model is further optimized with GRPO using format and outcome rewards.} 
\label{fig1}
\end{figure}

In this section, we present MediRec, a two-stage framework for explainable Chinese medication recommendation. As shown in Fig.~\ref{fig1}, MediRec is designed to simultaneously improve recommendation accuracy and reasoning interpretability by combining clinical reasoning distillation with reinforcement learning optimization. In the first stage, a powerful teacher LLM distills clinically grounded reasoning chains from real-world EHRs and drug instructions. The distilled reasoning data are then converted into supervised fine-tuning (SFT) samples to initialize the recommendation model. In the second stage, the model is further optimized with Group Relative Policy Optimization (GRPO) to improve the accuracy, consistency, and interpretability of generated medication recommendations.

\subsection{Stage 1: Clinical Reasoning Distillation and Supervised Fine-Tuning}

To enable fine-grained Chinese clinical medication recommendation, we establish a structured supervised fine-tuning (SFT) pipeline that synthesizes patient-specific records with generalized pharmacological knowledge. We first collect Chinese medication recommendation data from CHIP 2025 \cite{li2025overview}, which includes de-identified basic demographic characteristics, admission procedures, present illness, past medical history, chief complaint, discharge diagnoses, and discharge medications. To recommend medication more accurately, we also collect the corresponding drug package inserts from the DingXiangYuan website \footnote{https://www.dxy.cn/}, providing information on indications, usage, and precautions for each drug.

\begin{figure}
\includegraphics[width=\textwidth]{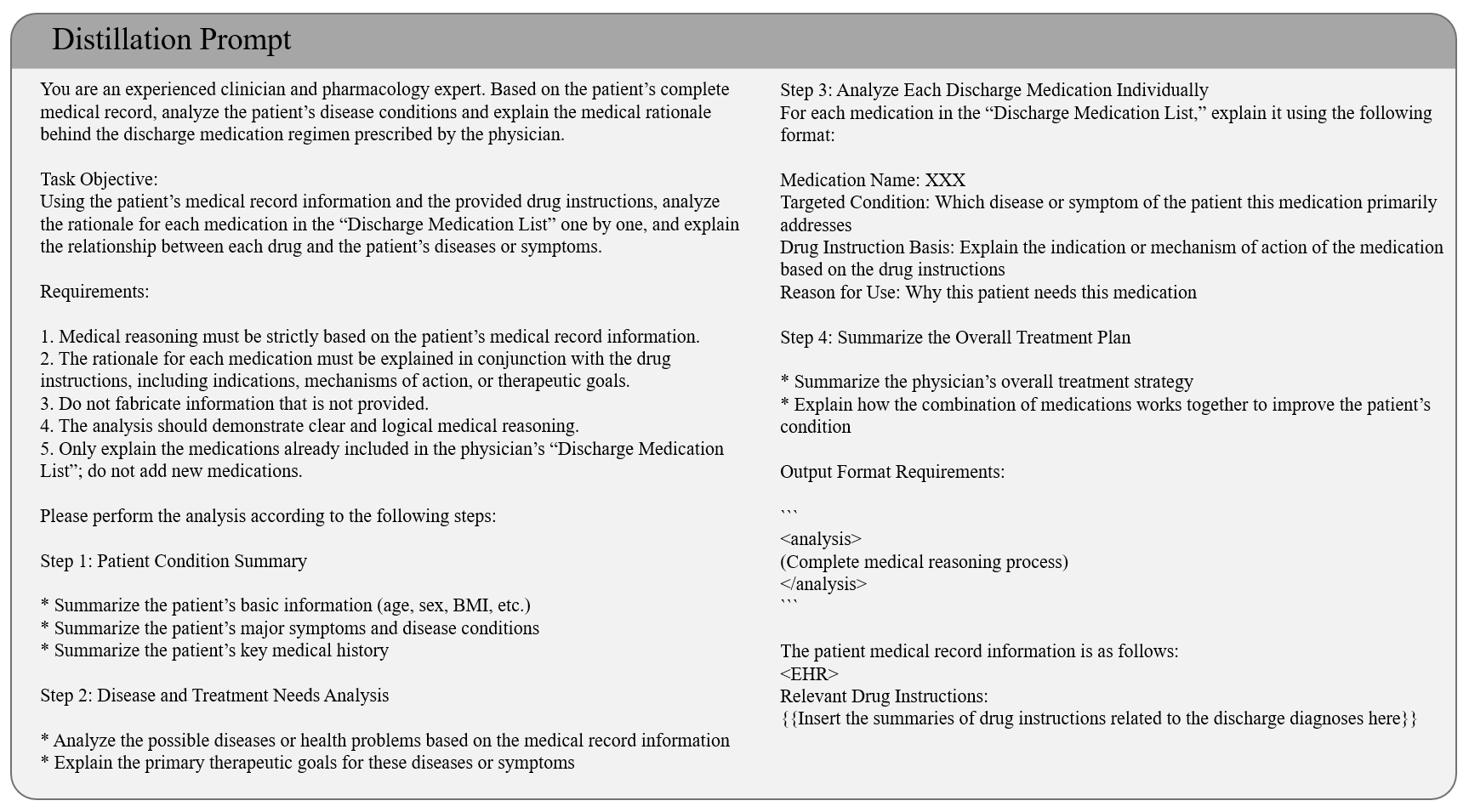}
\caption{Prompt template for clinical reasoning-chain distillation
} 
\label{fig2}
\end{figure}

Using these data sources, we employ a powerful teacher LLM (Qwen3-32B) to perform clinical reasoning-chain distillation. Given a patient’s EHR and the relevant drug instruction information, the teacher model generates interpretable reasoning trajectories consist of four major steps: (1) Patient Condition Summary: Summarizing the patient’s basic information, major symptoms, and key medical history; (2) Disease and Treatment Needs Analysis: Exploring potential diseases and explaining primary therapeutic goals based on the medical record; (3) Individual Medication Analysis: Providing per-drug explanations that align each prescribed medication with targeted conditions, mechanisms of action, and patient-specific needs; and (4) Overall Treatment Plan Synthesis: Summarizing the overarching treatment strategy and explaining how the combination of medications works synergistically. The distillation prompt is shown in Fig.~\ref{fig2}.

The second source is constructed from drug package inserts for CoT cold-start training. Structured drug labels containing information such as drug names, indications, contraindications, and pharmacological effects are collected and converted into question-answer (QA) pairs using predefined templates. For example, the ''Indications'' field of ''{Mekinist (Trametinib Tablets)'' is transformed into the question ``What are the indications of Mekinist (Trametinib Tablets)?'', while the corresponding label text serves as the answer. The same strategy is applied to other clinically relevant attributes, including contraindications and pharmacological effects. By converting structured pharmaceutical knowledge into explicit supervision signals, the model gains drug-specific supervision. Our tests show that fine-tuning with QA pairs improved Jaccard and F1 scores by 0.69\% and 0.63\%, respectively.

Following the distillation process, the reasoning trajectories and drug instruction QA pairs are converted into a standard SFT format. We then perform cold-start training using LoRA-based parameter-efficient fine-tuning. This dual-source data integration allows the model to jointly learn patient-specific clinical reasoning alongside drug-specific medical knowledge, equipping the policy model with the foundational clinical understanding required prior to reinforcement learning alignment.

\subsection{Stage 2: Reinforcement Learning via Policy Optimization}

Following SFT-based initialization, we further optimize the policy model using Group Relative Policy Optimization (GRPO). The objective of this stage is to improve the accuracy, robustness, and interpretability of medication recommendation generation. For GRPO training, we use only the reasoning-chain distillation data and exclude the drug instruction QA pairs, so that optimization focuses directly on patient-specific medication recommendation. Compared with standard supervised fine-tuning, GRPO enables the model to optimize generation quality directly against task-specific reward signals, effectively bridging the gap between token-level imitation and actual clinical recommendation utility.

To encourage the model to generate both structurally valid reasoning chains and accurate medication predictions, we design a hybrid reward function composed of two components: (1) \textit{Format reward}: A rule-based reward that evaluates whether the generated response strictly follows the predefined four-stage interpretable reasoning structure. (2) \textit{Outcome reward}: A performance-driven reward that measures the alignment between the predicted medications and the ground-truth medications (i.e., F1 and Jaccard metrics).

The final reward is defined as:
\[
R = \frac{1.5 \times \mathrm{F1} + 1.5 \times \mathrm{Jaccard} + R_{\mathrm{format}}}{4},
\]
where $\mathrm{F1}$ denotes the F1 score, $\mathrm{Jaccard}$ represents the Jaccard similarity coefficient, and $R_{\mathrm{format}}$ is the format reward score.

\section{Experiments and Results}
\subsection{Dataset and Experimental Settings}

We evaluate MediRec on the CHIP 2025 discharge medication recommendation benchmark, which contains de-identified EHRs collected from a tertiary hospital \cite{li2025overview}. Each record includes unstructured clinical text, such as the chief complaint, history of present illness, past medical history, admission condition, clinical course, and discharge diagnoses. The task is to recommend appropriate discharge medications from these EHRs. Table~\ref{tab1} summarizes the dataset statistics.

\begin{table}[]
\caption{The statistical information of the dataset.}\label{tab1}
\centering
\setlength{\tabcolsep}{6pt}
\begin{tabular}{lcccc}
\hline
Items                           & Train     & Val       & Test      & Total     \\ \hline
Number of Patients              & 1910      & 320       & 960       & 3190      \\
Number of Visit Records         & 3602      & 570       & 1722      & 5894      \\
Max/Min/Avg diagnosis & 23/1/6.82 & 21/1/6.85 & 24/1/6.79 & 24/1/6.82 \\
Max/Min/Avg drug       & 21/1/6.08 & 37/1/6.06 & 18/1/6.13 & 37/1/6.09 \\
diagnosis / drug space size     & 5363/586  & 1360/335  & 3093/459  & 7820/651  \\ \hline
\end{tabular}
\end{table}

The similarity coefficient (Jaccard), average precision (AVG\_P), recall (AVG\_R), and F1-score (F1) are used as evaluation metrics. The experiments are conducted on an NVIDIA A100 GPU. For SFT, the batch size is set to 1, gradient accumulation to 4, the learning rate to 1e-4, and the maximum number of epochs to 10. For GRPO training, the batch size is set to 1, gradient accumulation to 4, the learning rate to 5e-6, the number of generations to 8, and the maximum number of epochs to 8. The checkpoint with the highest validation F1 score is selected for final evaluation on the test set.

\subsection{Comparison of Prompting and Fine-Tuning Strategies}
To assess the intrinsic capabilities of LLMs for Chinese medication recommendation, we evaluate four open-source backbones: GLM4 (GLM4-9B-Chat), Llama3.1 (Llama3.1-8B-Instruct), Qwen3 (Qwen3-8B), and Baichuan-M1 (Baichuan-M1-14B). We compared zero-shot prompting (0-shot), one-shot prompting (1-shot), vanilla chain-of-thought prompting (CoT), and standard supervised fine-tuning (SFT). The results are shown in Fig.~\ref{fig3}. 

\begin{figure}
\includegraphics[width=\textwidth]{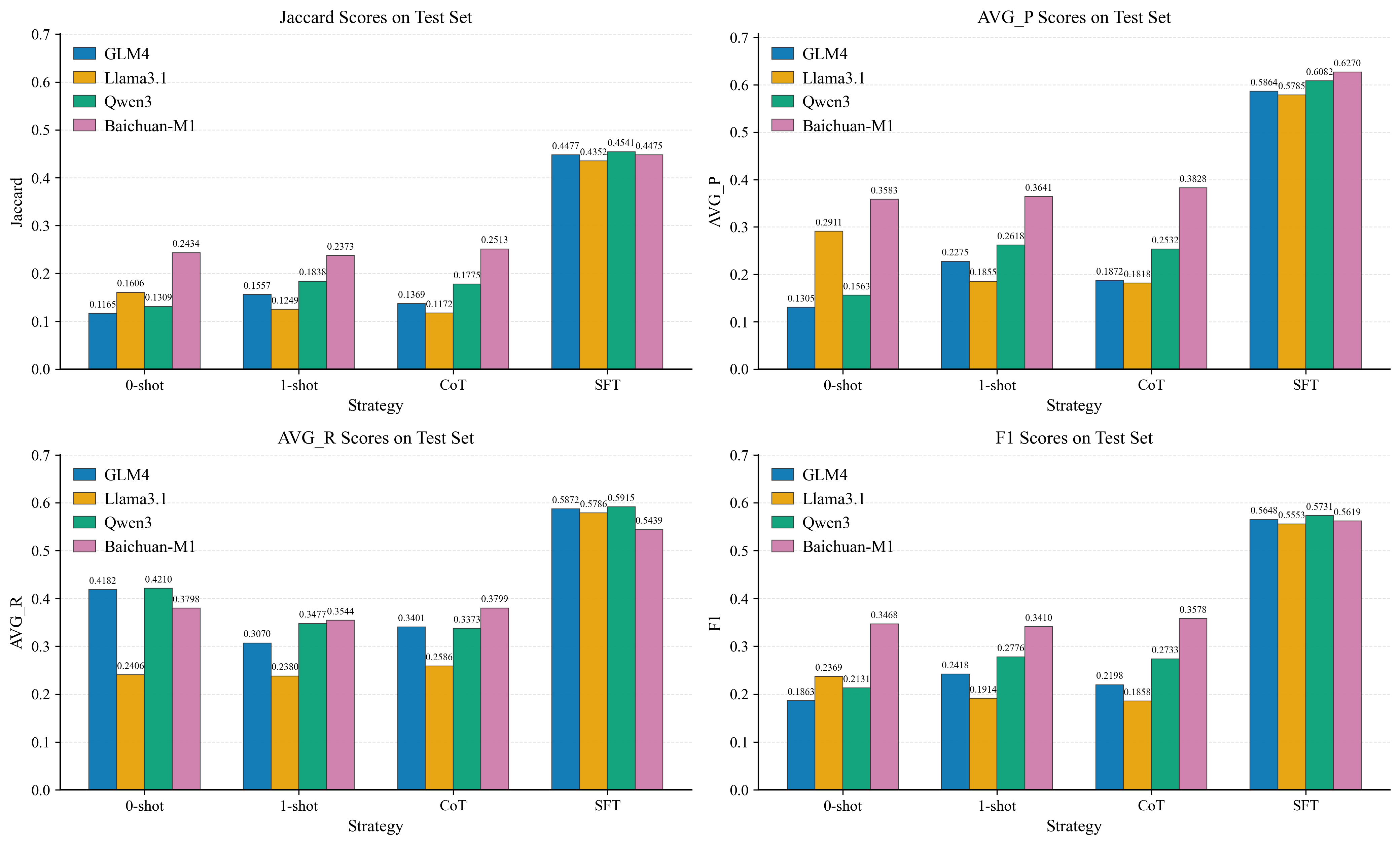}
\caption{Performance of different LLMs on the test set.} \label{fig3}
\end{figure}

From the results, some key findings emerge from this analysis: (1) The SFT strategy drastically outperforms all prompt-based strategies (0-shot, 1-shot, CoT) across every model family. This underscores that the intrinsic zero-shot reasoning of general LLMs cannot effectively handle fine-grained medication combinations without domain-specific training. (2) Providing a single example (1-shot) and vanilla CoT prompting yield minimal benefits over zero-shot prompting. This indicates that without explicit clinical alignment, unconstrained reasoning paths generate logical hallucinations that degrade final prediction accuracy. (3) Without SFT settings, domain-specific models (such as Baichuan-M1) show higher baseline stability due to integrated medical knowledge and larger model size. However, applying SFT drastically narrows this gap, allowing general-purpose backbones (like Qwen3) to achieve optimal performance through systematic instruction alignment.

\subsection{Effect of Reinforcement Learning}

We further evaluate the effect of reinforcement learning on two backbone models, GLM4 and Qwen3. For each backbone, we compare direct SFT, SFT with distilled clinical reasoning chains (CoT\_SFT), and GRPO optimization after reasoning-based cold-start training (CoT\_SFT-GRPO). The results are presented in Table~\ref{tab2}.

\begin{table}[]
\caption{Effect of Reinforcement Learning}\label{tab2}
\centering
\setlength{\tabcolsep}{6pt}
\begin{tabular}{lcccc}
\hline
Model       & Jaccard & AVG\_P & AVG\_R & F1     \\ \hline
Glm4-SFT        & 0.4477  & 0.5864 & 0.5872 & 0.5648 \\
Glm4-CoT\_SFT      & 0.4322  & 0.5694 & 0.5897 & 0.5530  \\
Glm4-CoT\_SFT-GRPO   & 0.4505  & 0.5924 & 0.5992 & 0.5709   \\
Qwen3-SFT    & 0.4541  & 0.6082 & 0.5915 & 0.5731   \\
Qwen3-CoT\_SFT  & 0.4400  & 0.5848 & 0.5890 & 0.5606  \\
Qwen3-CoT\_SFT-GRPO  & 0.4626  & 0.6139 & 0.6014 & 0.5813 \\ \hline
\end{tabular}
\end{table}

Introducing distilled reasoning chains (GLM4-CoT and Qwen3-CoT) slightly decreases direct prediction metrics compared with standard SFT. This suggests that reasoning-oriented supervision introduces a trade-off: the model learns to generate structured clinical explanations rather than only predicting medication labels. However, this stage provides a useful initialization for subsequent reinforcement learning. After applying GRPO with the hybrid reward function, both models achieve clear performance gains. Qwen3-CoT\_SFT-GRPO obtains the best result. These results demonstrate that GRPO effectively improves medication prediction while preserving the reasoning ability introduced during cold-start training.

We also conduct an ablation study on Qwen3 to examine the contribution of each reward component. As shown in Table~\ref{tab3}, removing any component reduces the F1 score. The format reward helps maintain structured and consistent reasoning, while the Jaccard and F1 rewards provide complementary supervision for multi-label medication prediction. The full reward design achieves the best overall balance, confirming the effectiveness of combining format control with outcome-based optimization.

\begin{table}[]
\caption{Effect of Reinforcement Learning and Reward Components}\label{tab3}
\centering
\setlength{\tabcolsep}{6pt}
\begin{tabular}{lcccc}
\hline
Model       & Jaccard & AVG\_P & AVG\_R & F1     \\ \hline
Qwen3-GRPO  & 0.4626  & 0.6139 & 0.6014 & 0.5813 \\
w/o Format  & 0.4643  & 0.6036 & 0.6009 & 0.5797 \\
w/o Jaccard & 0.4592  & 0.5886 & 0.6111 & 0.5772 \\
w/o F1      & 0.4608  & 0.6111 & 0.5916 & 0.5765 \\ \hline
\end{tabular}
\end{table}

\subsection{Performance Comparison with Existing Methods}

We compare Qwen3-based MediRec with standard fine-tuned LLMs and representative top-performing approaches from the CHIP 2025 challenge. To ensure a fair comparison, we consider only single-model configurations and exclude final multi-model ensemble results. The compared methods include (i) Zhu et al. \cite{zhu2025towards}, which uses a full-parameter fine-tuned Qwen3-8B model with pharmacological category augmentation and candidate-constrained post-processing; (ii) DP-EMR \cite{hua2025dp}, represented by its single ChatGLM4-9B model LoRA fine-tuned with data augmentation; and (iii) XiaoSu \cite{su2025lora}, which fine-tunes Qwen3-8B with LoRA and further improves it using pseudo-label-based retraining. The results are shown in Table~\ref{tab4}.

\begin{table}[]
\caption{Performance comparison with other existing methods on the test set.}\label{tab4}
\centering
\setlength{\tabcolsep}{6pt}
\begin{tabular}{lccccc}
\cline{1-5}
Model         & Jaccard & AVG\_P & AVG\_R & F1     &  \\ \cline{1-5}
Glm4          & 0.4477  & 0.5864 & 0.5872 & 0.5648 &  \\
Llama3.1      & 0.4352  & 0.5785 & 0.5786 & 0.5553 &  \\
Baichuan-M1   & 0.4475  & 0.6270 & 0.5439 & 0.5619 &  \\
Qwen3-8B      & 0.4541  & 0.6082 & 0.5915 & 0.5731 &  \\
Zhu et al. \cite{zhu2025towards}   & 0.4719  & 0.6194 & 0.6025 & 0.5885 & \\
DP-EMR \cite{hua2025dp}        & 0.4595  & 0.6165 & 0.5853 & 0.5749 & \\
XiaoSu \cite{su2025lora}    & 0.4324  & - & - & 0.5501 &  \\ 
MediRec (ours) & 0.4626  & 0.6139 & 0.6014 & 0.5813 &  \\
\cline{1-5}
\end{tabular}
\end{table}

By leveraging reinforcement learning, MediRec achieves superior performance compared to direct supervised fine-tuning (SFT), yielding results that are highly competitive with state-of-the-art approaches on the benchmark. While Zhu et al. achieve the highest Jaccard and F1 scores, their framework relies heavily on full-parameter fine-tuning and specialized post-processing strategies. In contrast, MediRec pairs its competitive predictive capabilities with an inherently explainable recommendation process. Unlike other methods that mainly output medication lists, MediRec generates clinical reasoning and supporting evidence for its recommendations, rendering it significantly more viable for practical clinical decision-support applications.

\subsection{Case Analysis}

To further validate the accuracy and interpretability of MediRec, an output example is shown in Fig.~\ref{fig4}, where the correct discharge medications are highlighted in red. 

\begin{figure}
\includegraphics[width=\textwidth]{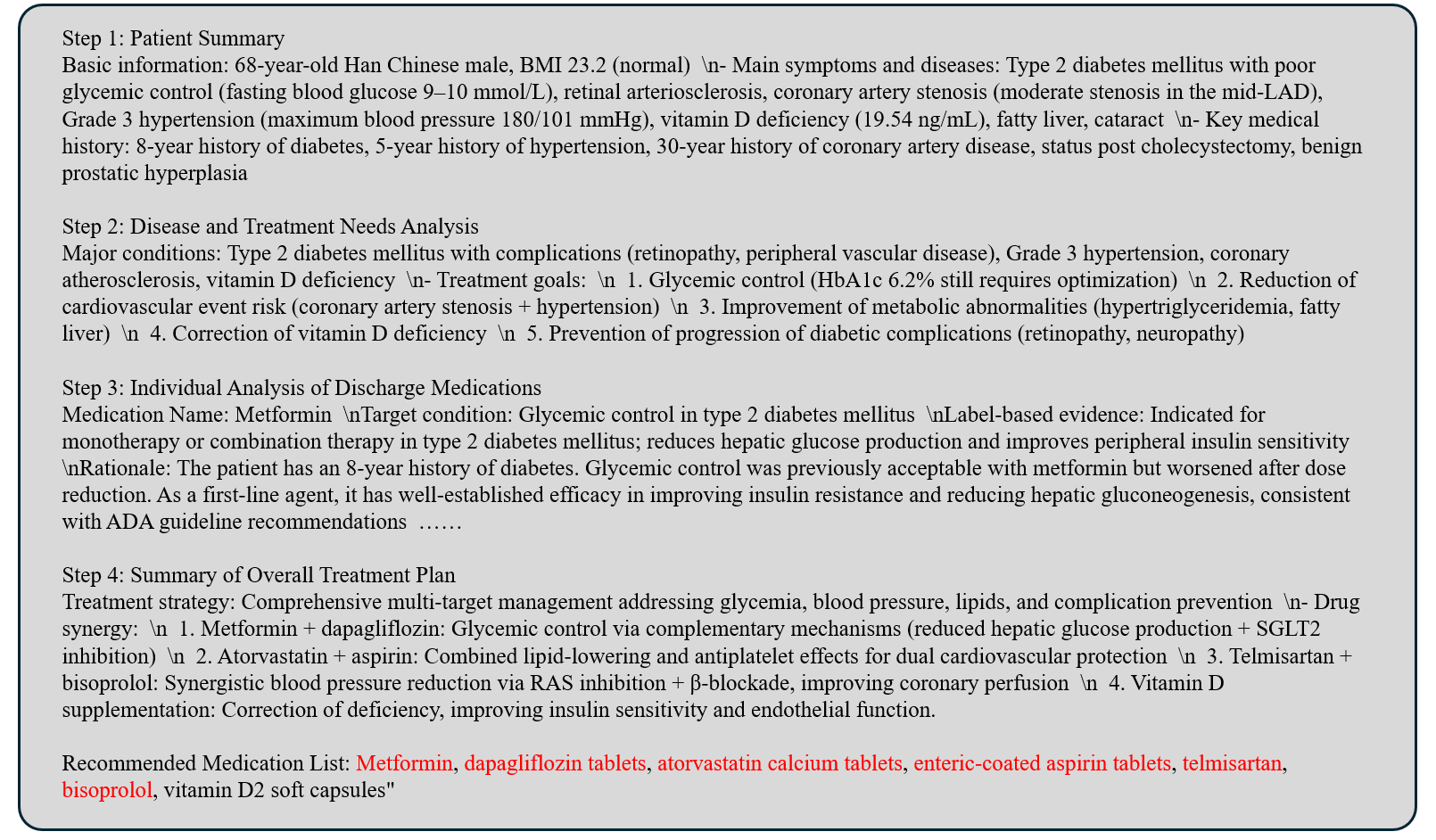}
\caption{Example of MediRec’s interpretable output.} \label{fig4}
\end{figure}

The case study shows that MediRec can align recommended medications with the patient’s clinical conditions and provide drug-level explanations based on therapeutic goals and pharmacological mechanisms. The generated recommendations are generally consistent with clinical treatment logic, especially for comprehensive cardiovascular risk management. However, the model still shows some limitations. It may over-recommend certain drugs, such as vitamin D supplementation, and some explanations may rely on broad clinical assumptions rather than strict evidence from the record. In addition, safety-related factors such as renal function and bleeding risk are not always explicitly discussed. These observations suggest that MediRec improves interpretability but still requires further refinement to enhance precision and safety awareness in clinical use.

\section{Conclusion}
In this work, we proposed MediRec, a two-stage framework for explainable Chinese discharge medication recommendation that combines clinical reasoning distillation with reinforcement learning. Experiments show that prompting alone is insufficient for this task, while domain-specific fine-tuning is essential. By integrating reasoning-based initialization and GRPO optimization, MediRec achieves competitive recommendation performance while generating interpretable clinical rationales, providing a more transparent alternative to methods that only predict medication lists.

Despite these promising results, several challenges remain. The model may still overlook certain safety considerations and occasionally over-recommend medications. In addition, the current framework relies mainly on EHR text and drug instructions. Future work will incorporate medical knowledge graphs, drug-drug interaction knowledge, and clinical guidelines to improve accuracy, safety, and interpretability.

\begin{credits}
\subsubsection{\ackname} This research was supported by the Natural Science Foundation of China (No. 62302076, 62276043), the Fundamental Research Funds for the Central Universities (No. DUT25YG108). 

\end{credits}
%
%
%
%




\bibliographystyle{plain}
\bibliography{references}

@article{liu2024large,
  title={Large language model distilling medication recommendation model},
  author={Liu, Qidong and Wu, Xian and Zhao, Xiangyu and Zhu, Yuanshao and Zhang, Zijian and Tian, Feng and Zheng, Yefeng},
  journal={arXiv preprint arXiv:2402.02803},
  year={2024}
}

@article{zhao2025addressing,
  title={Addressing overprescribing challenges: Fine-tuning large language models for medication recommendation tasks},
  author={Zhao, Zihao and Fan, Chenxiao and Gao, Chongming and Feng, Fuli and He, Xiangnan},
  journal={arXiv e-prints},
  pages={arXiv--2503},
  year={2025}
}

@article{fan2026fine,
  title={Fine-grained list-wise alignment for generative medication recommendation},
  author={Fan, Chenxiao and Gao, Chongming and Shi, Wentao and Gong, Yaxin and Zihao, Zhao and Feng, Fuli},
  journal={Advances in neural information processing systems},
  volume={38},
  pages={48037--48061},
  year={2026}
}

@article{johnson2016mimic,
  title={MIMIC-III, a freely accessible critical care database},
  author={Johnson, Alistair EW and Pollard, Tom J and Shen, Lu and Lehman, Li-wei H and Feng, Mengling and Ghassemi, Mohammad and Moody, Benjamin and Szolovits, Peter and Anthony Celi, Leo and Mark, Roger G},
  journal={Scientific data},
  volume={3},
  number={1},
  pages={1--9},
  year={2016},
  publisher={Nature Publishing Group}
}

@article{johnson2023mimic,
  title={MIMIC-IV, a freely accessible electronic health record dataset},
  author={Johnson, Alistair EW and Bulgarelli, Lucas and Shen, Lu and Gayles, Alvin and Shammout, Ayad and Horng, Steven and Pollard, Tom J and Hao, Sicheng and Moody, Benjamin and Gow, Brian and others},
  journal={Scientific data},
  volume={10},
  number={1},
  pages={1},
  year={2023},
  publisher={Nature Publishing Group UK London}
}

@article{grattafiori2024llama,
  title={The llama 3 herd of models},
  author={Grattafiori, Aaron and Dubey, Abhimanyu and Jauhri, Abhinav and Pandey, Abhinav and Kadian, Abhishek and Al-Dahle, Ahmad and Letman, Aiesha and Mathur, Akhil and Schelten, Alan and Vaughan, Alex and others},
  journal={arXiv preprint arXiv:2407.21783},
  year={2024}
}

@article{glm2024chatglm,
  title={Chatglm: A family of large language models from glm-130b to glm-4 all tools},
  author={Glm, Team and Zeng, Aohan and Xu, Bin and Wang, Bowen and Zhang, Chenhui and Yin, Da and Zhang, Dan and Rojas, Diego and Feng, Guanyu and Zhao, Hanlin and others},
  journal={arXiv preprint arXiv:2406.12793},
  year={2024}
}

@misc{qwen2025qwen25technicalreport,
      title={Qwen2.5 Technical Report}, 
      author={Qwen and : and An Yang and Baosong Yang and Beichen Zhang and Binyuan Hui and Bo Zheng and Bowen Yu and Chengyuan Li and Dayiheng Liu and Fei Huang and Haoran Wei and Huan Lin and Jian Yang and Jianhong Tu and Jianwei Zhang and Jianxin Yang and Jiaxi Yang and Jingren Zhou and Junyang Lin and Kai Dang and Keming Lu and Keqin Bao and Kexin Yang and Le Yu and Mei Li and Mingfeng Xue and Pei Zhang and Qin Zhu and Rui Men and Runji Lin and Tianhao Li and Tianyi Tang and Tingyu Xia and Xingzhang Ren and Xuancheng Ren and Yang Fan and Yang Su and Yichang Zhang and Yu Wan and Yuqiong Liu and Zeyu Cui and Zhenru Zhang and Zihan Qiu},
      year={2025},
      eprint={2412.15115},
      archivePrefix={arXiv},
      primaryClass={cs.CL},
      url={https://arxiv.org/abs/2412.15115}, 
}

@article{yang2021change,
  title={Change matters: Medication change prediction with recurrent residual networks},
  author={Yang, Chaoqi and Xiao, Cao and Glass, Lucas and Sun, Jimeng},
  journal={arXiv preprint arXiv:2105.01876},
  year={2021}
}

@article{wu2023dual,
  title={Dual attention and patient similarity network for drug recommendation},
  author={Wu, Jialun and Dong, Yuxin and Gao, Zeyu and Gong, Tieliang and Li, Chen},
  journal={Bioinformatics},
  volume={39},
  number={1},
  pages={btad003},
  year={2023},
  publisher={Oxford University Press}
}

@inproceedings{zhang2023hmfnet,
  title={E-HMFNet: A knowledge-enhanced hierarchical molecular representation fusion network for drug recommendation},
  author={Zhang, Junjie and Zang, Xuan and Chen, Hao and Tang, Buzhou},
  booktitle={2023 IEEE International Conference on Bioinformatics and Biomedicine (BIBM)},
  pages={1690--1695},
  year={2023},
  organization={IEEE}
}

@inproceedings{zhang2017leap,
  title={LEAP: learning to prescribe effective and safe treatment combinations for multimorbidity},
  author={Zhang, Yutao and Chen, Robert and Tang, Jie and Stewart, Walter F and Sun, Jimeng},
  booktitle={proceedings of the 23rd ACM SIGKDD international conference on knowledge Discovery and data Mining},
  pages={1315--1324},
  year={2017}
}

@article{gong2021smr,
  title={SMR: medical knowledge graph embedding for safe medicine recommendation},
  author={Gong, Fan and Wang, Meng and Wang, Haofen and Wang, Sen and Liu, Mengyue},
  journal={Big Data Research},
  volume={23},
  pages={100174},
  year={2021},
  publisher={Elsevier}
}

@article{read2011classifier,
  title={Classifier chains for multi-label classification},
  author={Read, Jesse and Pfahringer, Bernhard and Holmes, Geoff and Frank, Eibe},
  journal={Machine learning},
  volume={85},
  number={3},
  pages={333--359},
  year={2011},
  publisher={Springer}
}

@inproceedings{shang2019gamenet,
  title={Gamenet: Graph augmented memory networks for recommending medication combination},
  author={Shang, Junyuan and Xiao, Cao and Ma, Tengfei and Li, Hongyan and Sun, Jimeng},
  booktitle={proceedings of the AAAI Conference on Artificial Intelligence},
  volume={33},
  number={01},
  pages={1126--1133},
  year={2019}
}

@article{yang2021safedrug,
  title={Safedrug: Dual molecular graph encoders for recommending effective and safe drug combinations},
  author={Yang, Chaoqi and Xiao, Cao and Ma, Fenglong and Glass, Lucas and Sun, Jimeng},
  journal={arXiv preprint arXiv:2105.02711},
  year={2021}
}

@inproceedings{wu2022conditional,
  title={Conditional generation net for medication recommendation},
  author={Wu, Rui and Qiu, Zhaopeng and Jiang, Jiacheng and Qi, Guilin and Wu, Xian},
  booktitle={Proceedings of the ACM web conference 2022},
  pages={935--945},
  year={2022}
}

@article{hu2022lora,
  title={Lora: Low-rank adaptation of large language models.},
  author={Hu, Edward J and Shen, Yelong and Wallis, Phillip and Allen-Zhu, Zeyuan and Li, Yuanzhi and Wang, Shean and Wang, Liang and Chen, Weizhu and others},
  journal={Iclr},
  volume={1},
  number={2},
  pages={3},
  year={2022}
}

@article{dettmers2023qlora,
  title={Qlora: Efficient finetuning of quantized llms},
  author={Dettmers, Tim and Pagnoni, Artidoro and Holtzman, Ari and Zettlemoyer, Luke},
  journal={Advances in neural information processing systems},
  volume={36},
  pages={10088--10115},
  year={2023}
}

@article{zhang2025parameter,
  title={Parameter-efficient fine-tuning for foundation models},
  author={Zhang, Dan and Feng, Tao and Xue, Lilong and Wang, Yuandong and Dong, Yuxiao and Tang, Jie},
  journal={arXiv preprint arXiv:2501.13787},
  year={2025}
}

@article{luo2024taiyi,
  title={Taiyi: a bilingual fine-tuned large language model for diverse biomedical tasks},
  author={Luo, Ling and Ning, Jinzhong and Zhao, Yingwen and Wang, Zhijun and Ding, Zeyuan and Chen, Peng and Fu, Weiru and Han, Qinyu and Xu, Guangtao and Qiu, Yunzhi and others},
  journal={Journal of the American Medical Informatics Association},
  volume={31},
  number={9},
  pages={1865--1874},
  year={2024},
  publisher={Oxford University Press}
}

@article{dou2026baichuan,
  title={Baichuan-M3: Modeling Clinical Inquiry for Reliable Medical Decision-Making},
  author={Dou, Chengfeng and Yang, Fan and Li, Fei and Jia, Jiyuan and Ju, Qiang and Wang, Shuai and Li, Tianpeng and Zeng, Xiangrong and Zhou, Yijie and Zhang, Hongda and others},
  journal={arXiv preprint arXiv:2602.06570},
  year={2026}
}

@inproceedings{li2025overview,
  title={Overview of chip 2025 shared task 2: discharge medication recommendation for metabolic diseases based on Chinese electronic health records},
  author={Li, Juntao and Yuan, Haobin and Luo, Ling and Lv, Tengxiao and Jiang, Yan and Wang, Fan and Zhang, Ping and Lv, Huiyi and Wang, Jian and Sun, Yuanyuan and others},
  booktitle={China Health Information Processing Conference},
  pages={477--489},
  year={2025},
  organization={Springer}
}

@inproceedings{zhu2025towards,
  title={Towards Discharge Medication Recommendation via Multi-scale Model Training and Multi-dimensional Feature Enhancement},
  author={Zhu, Zhihong and Huang, Huimin and Wu, Xian},
  booktitle={China Health Information Processing Conference},
  pages={490--502},
  year={2025},
  organization={Springer}
}

@inproceedings{hua2025dp,
  title={DP-EMR: A Chinese Medication Recommendation Method for Metabolic Diseases Based on Two-Stage Ensemble Learning},
  author={Hua, Zhongtian and Wu, Kejun and Ru, Chengxin and Luo, Yi and Wang, Mengyuan and Yu, Meijia and Han, Yingjie},
  booktitle={China Health Information Processing Conference},
  pages={503--517},
  year={2025},
  organization={Springer}
}

@inproceedings{su2025lora,
  title={LoRA-Fine-Tuned LLMs for Discharge Medication Recommendation on Chinese EHRs},
  author={Su, Xiao},
  booktitle={China Health Information Processing Conference},
  pages={518--530},
  year={2025},
  organization={Springer}
}
\end{document}